\documentclass[final]{cvpr}
\usepackage{times}
\usepackage{epsfig}
\usepackage{graphicx}
\usepackage{amsmath}
\usepackage{amssymb}

\usepackage{multirow}
\usepackage{subfigure}
\usepackage{tabularx}
\usepackage{color}
\usepackage{rotating}
\usepackage{diagbox}
\usepackage{bm}

\usepackage[pagebackref=true,breaklinks=true,colorlinks,bookmarks=false]{hyperref}

\def\confYear{CVPR 2021}

\begin{document}
\title{Unbalanced Feature Transport for Exemplar-based Image Translation}


\author{
Fangneng Zhan \textsuperscript{\rm 1,2},
Yingchen Yu \textsuperscript{\rm 1,2},
Kaiwen Cui \textsuperscript{\rm 1},
Gongjie Zhang \textsuperscript{\rm 1}, 
Shijian Lu \textsuperscript{\rm 1}, 
Jianxiong Pan \textsuperscript{\rm 2}, \\  
Changgong Zhang \textsuperscript{\rm 2},
Feiying Ma \textsuperscript{\rm 2},
Xuansong Xie \textsuperscript{\rm 2}, 
Chunyan Miao \textsuperscript{\rm 1} \\
\textsuperscript{\rm 1} Nanyang Technological University \quad  \textsuperscript{\rm 2} DAMO Academy, Alibaba Group
}

\maketitle

\thispagestyle{empty}
\pagestyle{empty}

\begin{abstract}
Despite the great success of GANs in images translation with different conditioned inputs such as semantic segmentation and edge maps, generating high-fidelity realistic images with reference styles remains a grand challenge in conditional image-to-image translation. This paper presents a general image translation framework that incorporates optimal transport for feature alignment between conditional inputs and style exemplars in image translation. The introduction of optimal transport mitigates the constraint of many-to-one feature matching significantly while building up accurate semantic correspondences between conditional inputs and exemplars. We design a novel unbalanced optimal transport to address the transport between features with deviational distributions which exists widely between conditional inputs and exemplars. In addition, we design a semantic-activation normalization scheme that injects style features of exemplars into the image translation process successfully. Extensive experiments over multiple image translation tasks show that our method achieves superior image translation qualitatively and quantitatively as compared with the state-of-the-art.
\end{abstract}

\section{Introduction}

Conditional image-to-image translation aims to generate images from certain given conditional inputs such as semantic segmentation \cite{park2019spade,wang2018pix2pixhd}, layout \cite{li2020bachgan}, and key points \cite{tang2019cycle}.
With the advance of Generative Adversarial Networks (GANs), it has made rapid progress and achieved quite promising translation performance in recent years.
However, most existing methods have very loose control over the translation process which often affects the translation quality greatly and so the wide application of image translation in various tasks. Optimal style control is still an open challenge in high-fidelity realistic image translation.

\begin{figure}[t]
\centering
\includegraphics[width=1.0\linewidth]{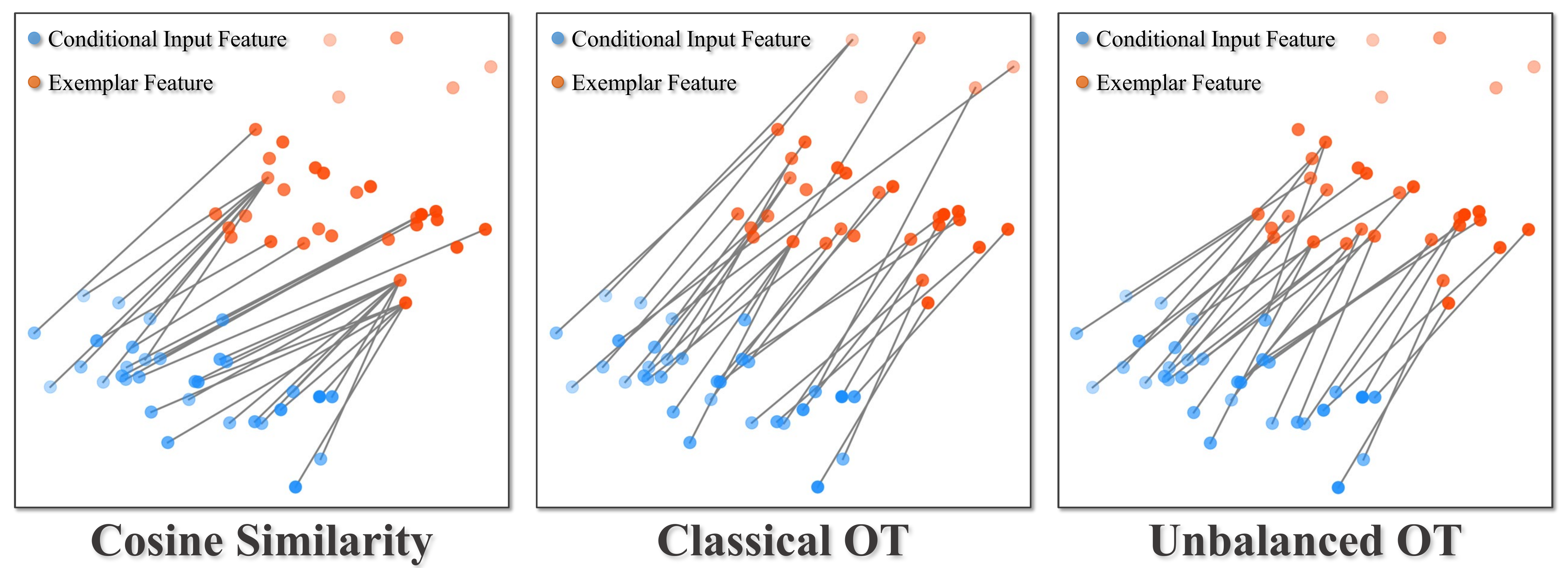}
\caption{
Different feature matching in image translation: \textit{Cosine Similarity} tends to match each feature separately which often leads to many-to-one matching. Classical optimal transport (\textit{Classical OT}) suppresses the many-to-one matching problem but it matches all feature points including undesired outliers (existing between deviational feature distributions). Our designed unbalanced optimal transport (\textit{Unbalanced OT}) mitigates many-to-one matching and avoid outlier matching effectively.
}
\label{im_intro}
\end{figure}

Several prior works attempted to tackle the style control challenge by using a latent code that is encoded by either Variational Auto-Encoder (VAE) \cite{park2019spade} or style encoder \cite{choi2020starganv2}.
However, latent codes often impair style control accuracy as they do not have sufficient capacity to capture detailed style information. 
A different approach is to inject specific style codes for different semantic regions \cite{zhu2020sean}, but it is specifically designed for conditional input of semantic segmentation and cannot well generalize to other conditional inputs. Recently, Zhang et al.  \cite{zhang2020cocosnet} explore to establish dense semantic correspondence between conditioned input and a given style exemplar so as to offer dense style guidance in translation.
However, it constructs the semantic correspondence based on cosine similarity that often leads to many-to-one matching (i.e. multiple conditional input features match to the same exemplar feature as illustrated in Fig. \ref{im_intro}) and missing of details in image translation.

We designed \textit{UNITE}, UNbalanced optImal feature Transport for Exemplar-based image translation that achieves high-fidelity image generation with faithful style to given exemplars. UNITE consists of a \textit{feature transport network} and a \textit{translation network} that are inter-connected and can be jointly optimized in training. The \textit{feature transport network} introduces optimal transport \cite{peyre2019computational} which matches two sets of features as a whole and effectively overcomes many-to-one matching as in the widely adopted cosine similarity \cite{zhang2020cocosnet} that matches individual features separately. To tackle the distribution deviations between conditional inputs and exemplars, we design an unbalanced optimal transport technique that adaptively learns the mass (or weight) of each individual feature for effective transport between distributions of different masses. In the \textit{translation network}, we design a semantic-activation normalization scheme that injects the aligned features into the translation process, where the exemplar features are transported in a multi-stage manner for preserving rich and complicated textural details. Extensive experiments show that UNITE translates images with superior realism and fidelity. 

The contributions of this work can be summarized in three aspects. First, we propose a conditional image translation framework that introduces optimal transport for proper feature alignment and faithful style control in image translation. Second, we design an unbalanced optimal transport technique with adaptive mass learning scheme that is capable of aligning features with deviational distributions, and a multi-stage transport strategy that can preserve complex textures at different scales. Third, we design a novel semantic-activation normalization that is capable of injecting the aligned style features into the image translation process effectively.

\section{Related Work}

\subsection{Image-to-Image Translation}

GAN-based image-to-image translation has been investigated extensively due to its wide applications in different tasks such as domain adaptation \cite{shrivastava2017simgan,zhan2019gadan,zhan2019esir,zhan2020sagan},
data augmentation \cite{zhang2021defect,zhan2018verisimilar,zhan2019scene,xue2018accurate},
image editing \cite{yu2018inpainting,wu2020cascade,wu2020leed,koksal2020rf,yu2021diverse}, image composition \cite{zhan2019sfgan,zhan2019acgan,zhan2020aicnet,zhan2020emlight,zhan2021gmlight,zhan2021rabit,zhan2020towards}, etc. Existing works explored different conditional inputs such as semantic segmentation \cite{isola2017pix2pix,wang2018pix2pixhd,park2019spade}, scene layouts \cite{sun2019lostgan,zhao2019layout2im,li2020bachgan}, key points \cite{ma2017pose,men2020adgan,zhang2021deep}, edge maps \cite{isola2017pix2pix,zhu2017toward,lee2018diverse}, etc. for photo-realistic image translation. On the other hand, optimal style control remains a critical yet challenging task that has attracted increasing attention in recent years. For example, \cite{huang2018multimodal} and \cite{ma2018exemplar} transfer style codes from exemplars to source images via adaptive instance normalization (AdaIN) \cite{huang2017adain}. \cite{park2019spade} uses variational autoencoder (VAE) \cite{kingma2013vae} to encode exemplars for image translation. \cite{choi2020starganv2} employs a style encoder for style consistency between exemplars and the translated images. 

Different from the aforementioned methods that adopt latent vectors for style control, \cite{zhang2020cocosnet} learns dense semantic correspondences between conditional inputs and exemplars for image translation. Similar ideas have been explored in other translation tasks such as image colorization \cite{he2018colorization,zhang2019colorization} that also employs exemplars to build up semantic correspondences. On the other hand, most existing works use cosine similarity to build up semantic correspondences which often suffer from many-to-one matching and resultant feature missing. We introduce optimal transport for feature matching that treats the whole feature set as a whole and overcomes the many-to-one matching effectively.

\subsection{Optimal Transport}

Optimal transport (OT) \cite{villani2008ot} provides a principal way of comparing distributions and offers optimal plans for matching distributions. As a linear programming problem, classic OT is computationally intensive and \cite{cuturi2013sinkhorn} presents entropy regularized optimal transport that is differentiable and can be solved by the Sinkhorn-Knopp algorithm \cite{sinkhorn1967concerning,knight2008sinkhorn} efficiently. On the other hand, the classical optimal transport has a typical constraint that they can only handle distributions with equal mass and thus become inapplicable while facing unbalanced distributions with different masses and deviations as widely existed in various tasks. Different approaches have been reported to address this new challenge. For example, \cite{chizat2016uot} presents a unified treatment of unbalanced optimal transport that allows for both static and dynamic formulations. \cite{liero2015uot} introduce an entropic version of unbalanced optimal transport.

In recent years, optimal transport has been widely explored in various computer vision tasks such as domain adaptation \cite{courty2016optimal}, semantic matching \cite{liu2020scot}, style transfer \cite{kolkin2019style}, etc. In this work, we adapt unbalanced feature transport for aligning deviational features between conditional inputs and exemplars for high-fidelity image translation.

\begin{figure*}[t]
\centering
\includegraphics[width=1.0\linewidth]{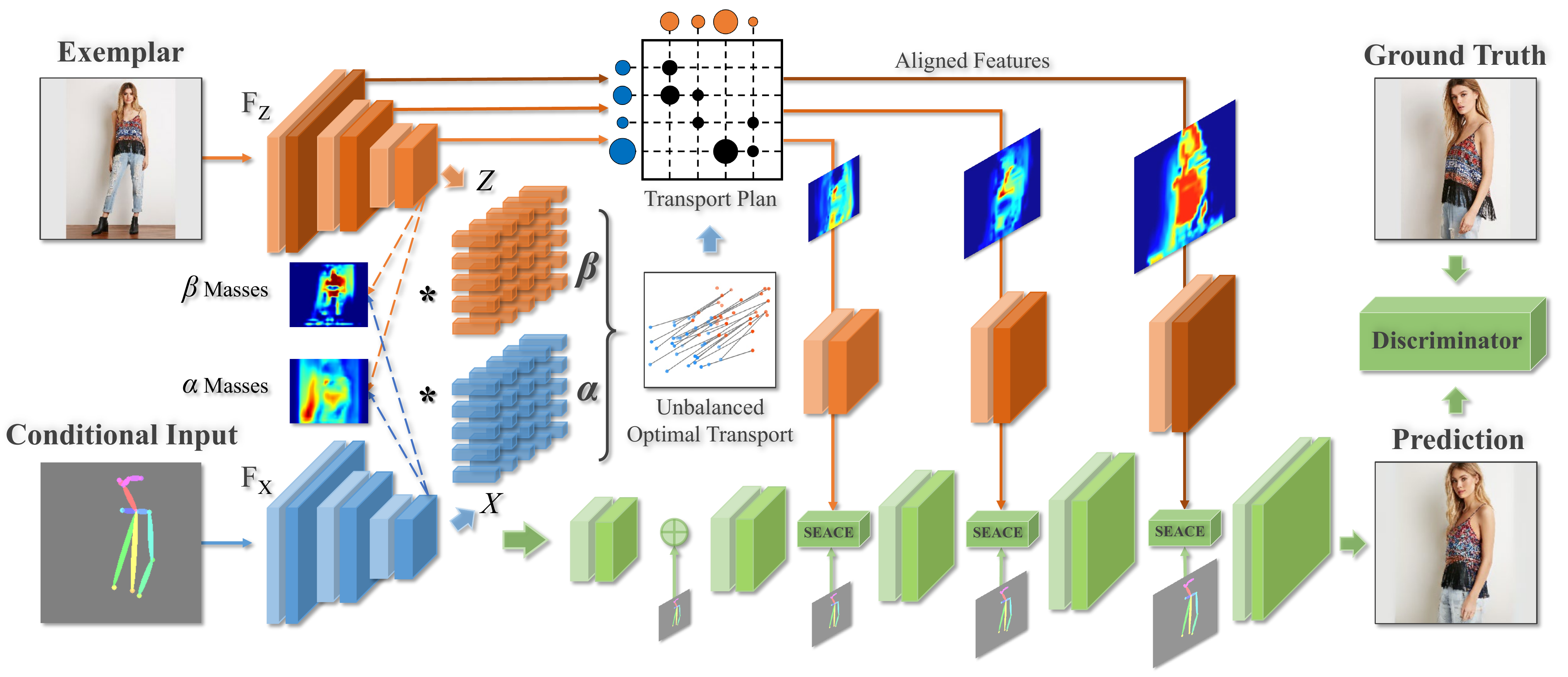}
\caption{
The framework of our proposed network: The \textit{Conditional Input} and \textit{Exemplar} are fed to feature extractors $F_{X}$ and $F_{Z}$ to extract feature vectors $X$ and $Z$. The mass (or weight) of the feature vectors ($\alpha$ and $\beta$ masses) are then determined collectively by $X$ and $Z$. 
The weights and the feature vectors form two sets of Dirac masses $\bm{\alpha}$ and $\bm{\beta}$, which are further aligned through \textit{Unbalanced Optimal Transport}. With an obtained \textit{Transport Plan}, the feature of the \textit{Exemplar} is transported in a multi-stage manner to be aligned with that of the \textit{Conditional Input}. The aligned features will be injected into the translation network through a proposed SEmantic-ACtivation (dE)normalization (SEACE) to synthesize the final output image. (Blue and orange parts for feature transport network, green part for translation network)}
\label{im_stru}
\end{figure*}

\section{Proposed Method}
Our UNITE consists of a feature transport network (in blue and orange) and a translation network (in green) which are inter-connected as shown in Fig. \ref{im_stru}. The feature transport network aligns the features of conditional inputs and exemplars and the translation network produces the final synthesis, more details to be described in the following subsections.

\subsection{Feature Transport Network}

The feature transport network aims to transport the feature of exemplars to be aligned with that of conditional inputs, thus providing accurate style guidance for the image translation. As shown in Fig. \ref{im_stru}, both conditional input and exemplar are fed to two feature extractors $F_{X}$ and $F_{Z}$ to extract two sets of feature vectors $X = (x_1, \cdots, x_{n}) \in \mathbb{R}^{d}$ and $Z = (z_1,\cdots,z_{n}) \in \mathbb{R}^d$, where $n$ denotes the number of feature vectors and $d$ denotes the feature dimension.

\begin{figure}[t]
\centering
\includegraphics[width=1.0\linewidth]{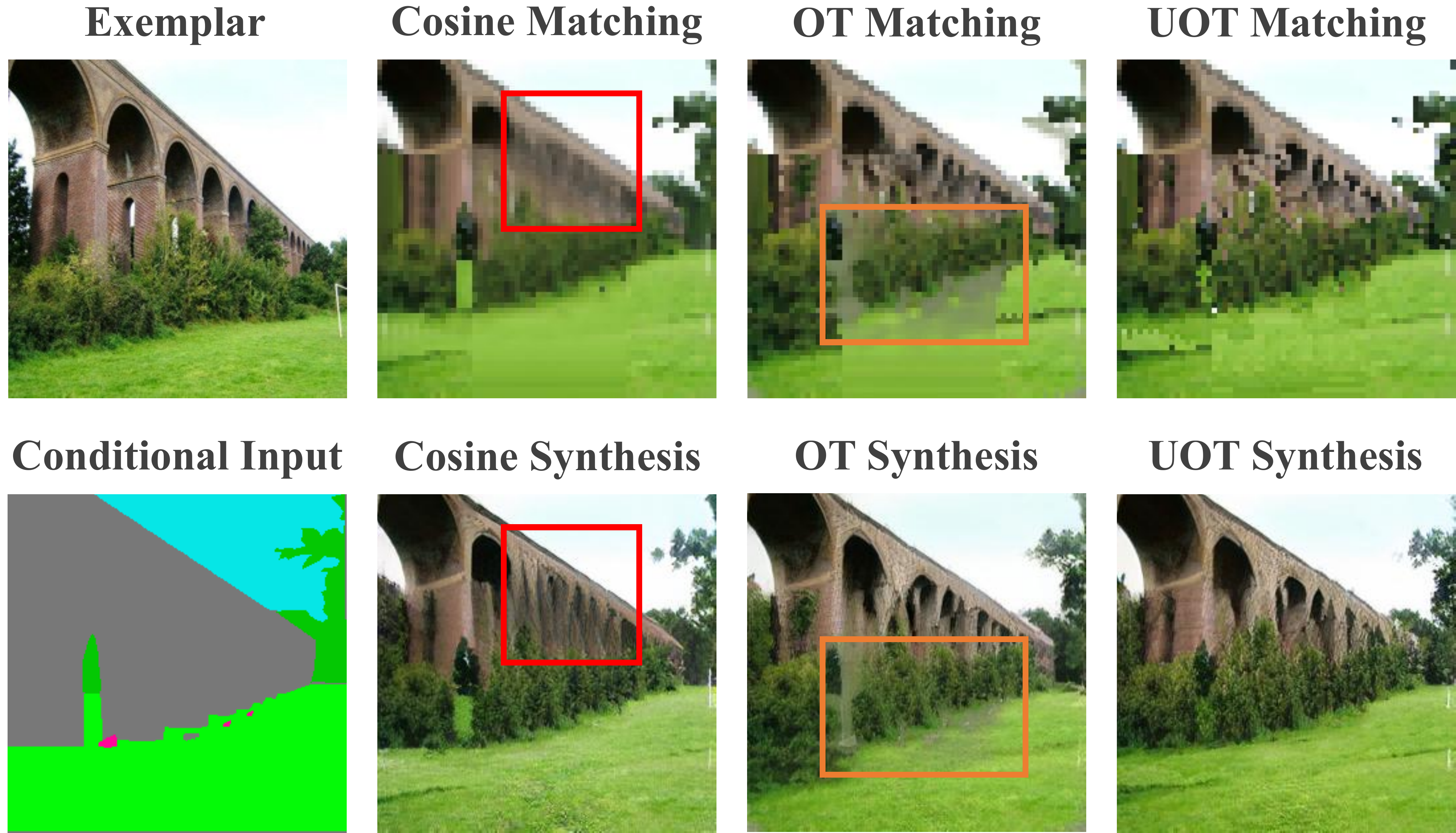}
\caption{
The comparison of different feature alignment methods: 
For visual comparison, we directly apply the feature alignment result to warp the exemplar.
The \textit{Cosine Matching} using cosine similarity often leads to many-to-one matching that introduces blurry feature alignment as highlighted by red box, which further leads to blurry synthesis result as shown in \textit{Cosine Synthesis}. \textit{OT Matching} using classical optimal transport suppresses the many-to-one matching but tends to introduce false matching as highlighted by orange box. Our proposed \textit{UOT Matching} using unbalanced optimal transport mitigates both many-to-one matching and false matching effectively, which achieve the best feature alignment and synthesis fidelity as illustrated in \textit{UOT Synthesis}.
}
\label{im_align}
\end{figure}

To align feature sets $X$ and $Z$, most existing methods \cite{zhang2020cocosnet,he2018colorization,zhang2019colorization} build a dense correspondence matrix between $X$ and $Z$ by measuring the Cosine similarity between any two feature vectors. As each feature vector $x_{i}$ is matched to the feature vector $z_{j}$ with the maximum Cosine similarity separately, multiple feature vectors in $X$ may correspond to the same feature vector in $Z$ (i.e. many-to-one matching), which leads to blurry translation as illustrated in Fig. \ref{im_align}. To avoid many-to-one matching between sets of feature vectors, we introduce optimal transport method to align the features of conditional inputs and exemplars.

\textbf{Classical Optimal Transport.}
The classical Optimal transport aims to determine the best transport plan (namely the minimum amount of total work required) to transform one measure into another with the same mass. Here the `work’ is evaluated by the product of the cost and the amount of mass to be transported. With constraints on the total masses in transport, optimal transport penalizes the many-to-one matching effectively.

To formulate the feature alignment as an optimal transport problem and derive the constraints of total masses, we encode the conditional input feature $X$ and exemplar feature $Z$ as Dirac masses: $\alpha = \sum_{i=1}^{n} \alpha_{i} \delta_{x_{i}}$ and  $\beta = \sum_{i=1}^{n} \beta_{i} \delta_{z_{j}}$, where the masses $\alpha_{i}, \beta_{i} \geq 0$ and feature vectors $x_{i},z_{i}$ denote the locations of $\alpha_{i},\beta_{i}$. Then we define a distance matrix $C$, where each entry $C_{ij}$ in $C$ gives the cost of moving mass $\alpha_{i}$ to mass $\beta_{j}$ which can be defined by:
$
C_{ij} = 1 - \frac{x_{i}^{\top} \cdot z_{j}} {||x_{i}|| \, ||z_{j}||}
$
A transport plan $T$ can be defined, where each entry $T_{ij}$ is the amount of masses transported between $\alpha_{i}$ and $\beta_{j}$. Then the classical optimal transport problem can be formed as:
\begin{equation}
\begin{split}
& OT(\alpha,\beta) = \mathop{\min}\limits_{T} (\sum_{i,j=1}^{n} C_{ij} T_{ij}) = \mathop{\min}\limits_{T} \langle C, T \rangle \\
&{\rm subject \ to} \quad (T \vec{1}) = \alpha, \quad (T^\top \vec{1}) = \beta \\ 
\end{split}
\label{formula_ot}
\end{equation}
The constraints of the total masses $\quad (T \vec{1}) = \alpha$ and $(T^\top \vec{1}) = \beta$ naturally penalize the many-to-one matching in optimal transport as illustrated in Fig. \ref{im_align}.

\textbf{Unbalanced Optimal Transport.}
For classical optimal transport, the total masses of the two measures should be the same, namely $\sum_{i=1}^{n} \alpha_{i} = \sum_{j=1}^{n} \beta_{j}$. But for conditional inputs and exemplars, their features are usually not perfectly matched so have different total masses.
For example, the conditional input (key-point map) in Fig. \ref{im_stru} does not contain feet which exist in the exemplar, so the feature of feet region in the exemplar is treated as outliers in optimal transport and should not be matched to any feature of the conditional input.
However, classical optimal transport inevitably matches all features, leading to inaccurate or false matching as illustrated in Fig. \ref{im_align}.
We handle it by introducing a relaxed version of classical optimal transport, namely unbalanced optimal transport (UOT or unbalanced OT) \cite{chizat2016uot} that aims to determine an optimal transport plan between measures of different total masses. We formulate unbalanced OT by replacing the `Hard' conservation of masses in (\ref{formula_ot}) by a `Soft' penalty with a divergence metric. An unbalanced OT problem can thus be formulated as follows:
\begin{equation}
\begin{split}
 & \mathop{\min}\limits_{T} \left [ \langle C, T \rangle + \tau {\rm KL} (T \vec{1} || \alpha) + \tau {\rm KL} (T^{\top} \vec{1}|| \beta) \right ]  \\
\end{split}
\end{equation}
where $\tau$ is regularization parameter, ${\rm KL}$ is the Kullback-Leibler divergence which is defined as ${\rm KL} (a||b) = \sum_{i = 1}^{n} a_{i} \log (\frac{a_{i}}{b_{i}}) - a_{i} + b_{i}$.

We employ cross-inner product to generate the masses $\alpha_{i}, \beta_{j} (i,j \in [1,n])$ associated with each feature vector. The masses are highly correlated with specific conditional inputs and exemplars, thus it should be determined collectively by both of them. Intuitively, the feature vector that is more related with another feature set should have higher mass. We therefore determine the mass of a feature vector by computing its relevance with another feature set:
\begin{equation}
    \alpha_{i} = x_{i}  \cdot \frac{\sum_{i=1}^{n}(z_{i})}{n}, \;
    \beta_{j} = z_{j}  \cdot \frac{\sum_{j=1}^{n}(x_{j})}{n}
\end{equation}
The mass parameters are adaptively updated in training. They capture the mass of each single feature vector accurately and mitigate the false matching problem effectively.

To implement UOT in a differentiable manner, an entropic regularization term $H(T) = - \sum_{i,j=1}^{n} T_{ij} \log T_{ij}$ is introduced. An entropic UOT problem can be defined by:
\begin{equation*}
\begin{split}
 & \mathop{\min}\limits_{T} \left [ \langle C, T \rangle + \tau {\rm KL} (T \vec{1} || \alpha) + \tau {\rm KL} (T^{\top} \vec{1}|| \beta) - \eta H(T)  \right ]  \\
\end{split}
\end{equation*}
where $\eta$ is the regularization coefficients that denotes the smoothness of the transport plan $T$. In our network, $\eta$ is fixed at 0.0001 empirically.

To obtain $T$, we consider the Fenchel-Legendre dual form of the entropic UOT that is defined by:
\begin{footnotesize}
\begin{equation}
\begin{split}
& \mathop{\max}\limits_{u, v} \left[ -F^*(-u) - G^*(-v) - \eta \sum_{i,j}\exp(\frac{u_i + v_j - C_{ij}}{\eta})
\right] \\
\label{formula_dual}
\end{split}
\end{equation}
\end{footnotesize}
where $F^*$ and $G^*$ are the Legendre conjugate of ${\rm KL}$ divergence which can be computed by:
\begin{equation*}
\begin{split}
& F^*(u) = \mathop{\max}\limits_{z} z^{\top}u - \tau {\rm KL}(z||\alpha) = \tau \langle e^{u/\tau}, \alpha \rangle - \alpha^{\top} \vec{1} \\
& G^*(v) = \mathop{\max}\limits_{x} x^{\top}v - \tau {\rm KL} (x||\beta) = \tau \langle e^{v/\tau}, \beta \rangle - \beta^{\top} \vec{1} \\
\end{split}
\end{equation*}
Then the Sinkhorn algorithm \cite{cuturi2013sinkhorn} can be applied to (\ref{formula_dual}) for approximating UOT solution, with a desired transport plan $T$ encoded by optimal dual vectors $u$ and $v$ as below:
\begin{equation}
    T_{ij} = \alpha_{i} \beta_{j} \, \exp \frac{1}{\eta} \left [ u_{i} + v_{j} - C_{ij} \right]
\end{equation}

\textbf{Multi-Stage Feature Transport}.
With the transport plan, the exemplar features can be transported to be aligned with conditional input features for translation. Different from CoCosNet \cite{zhang2020cocosnet} that warps exemplar images directly, we adopt a multi-stage manner to transport exemplar features as shown in Fig. \ref{im_stru}. This multi-stage transport helps to preserve detailed exemplar features especially for textures with complicated patterns as illustrated in Fig. \ref{im_com}.

\subsection{Translation Network}

The translation network aims to synthesize images under the semantic guidance of conditional inputs and style guidance of aligned exemplar features. The overall architecture of the translation network is similar to SPADE \cite{park2019spade} as illustrated in Fig. \ref{im_stru} (green part). More details of the network structure are available in the supplementary material.

In translation network, the aligned exemplar features are injected into the generation process at multiple stages to control the style of output image. Although style feature injection can be handled by several different approaches such as SPADE \cite{park2019spade}, all prevalent approaches fail to consider the semantic correlation between style features in feature injection. We designed an innovative semantic-aware injection method to be described in the following subsection.

\begin{figure}[t]
\centering
\includegraphics[width=1.0\linewidth]{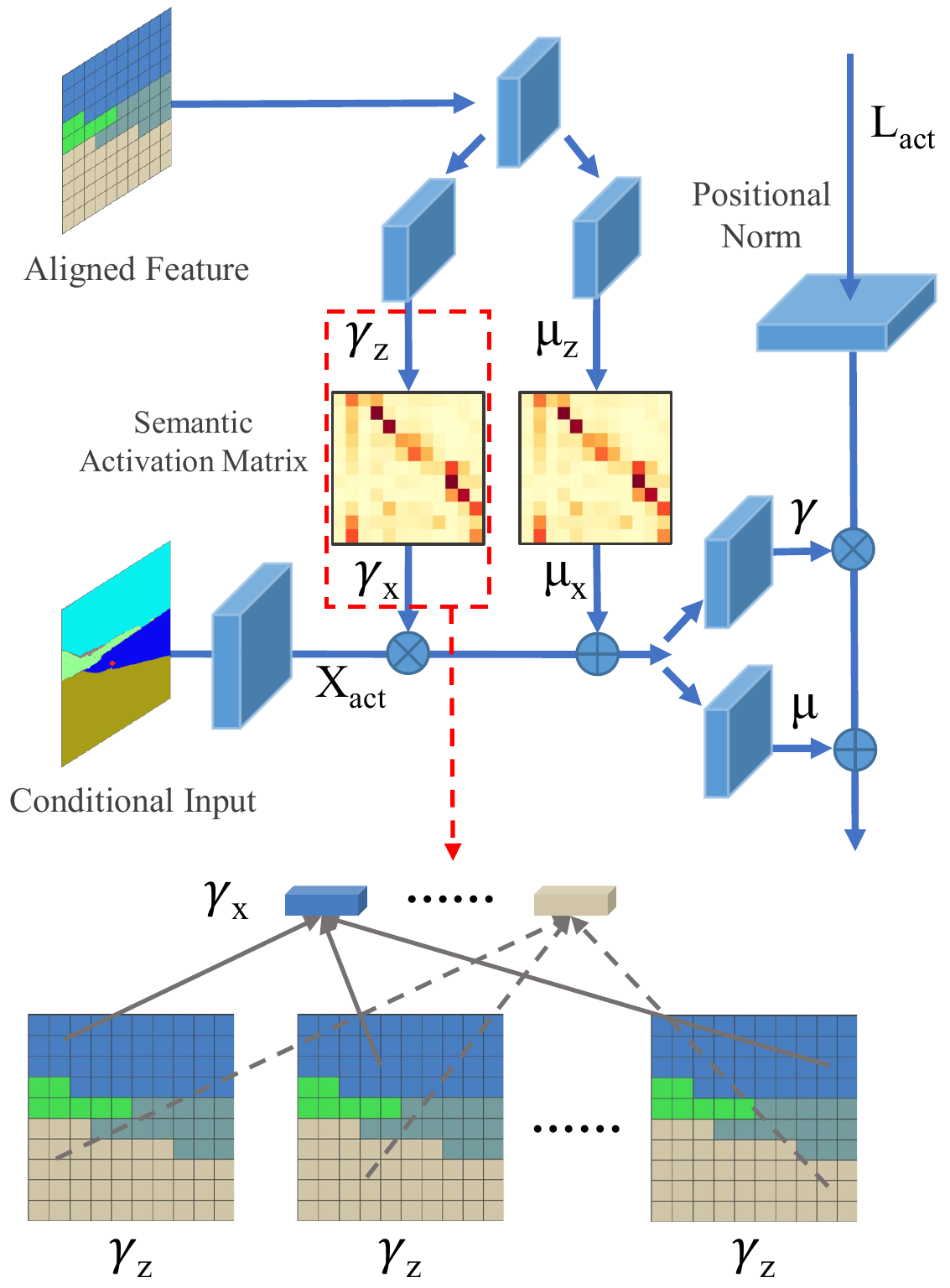}
\caption{
The structure of the proposed SEmantic-ACtivation (dE)normalization (SEACE): To build the long range dependency between style features, a semantic-activation matrix is obtained by computing the self-attention of the condition input features $X$ that are extracted in the feature transport network. With the semantic-activation matrix, $\gamma_{\scriptscriptstyle X}$ is determined collectively by the entire region in $\gamma_{\scriptscriptstyle Z}$ with the same semantic as shown at the bottom.
}
\label{im_norm}
\end{figure}

\textbf{Semantic-Activation Denormalization.}
Ideally, the style of a spatial position should be determined by all the style feature with the same semantic instead of only relying on the local feature in the exemplar. In addition, building long-range dependencies between style features is usually beneficial to image generation \cite{zhang2019self} as it allows to leverage the complementary style features of distant image regions. Based on these observations, we propose a novel SEmantic-ACtivation (dE)normalization (SEACE) to model the long-range dependencies across style features in style injection.

As shown in Fig. \ref{im_norm}, two sets of modulation parameters $\gamma_{\scriptscriptstyle Z}$ and $\mu_{\scriptscriptstyle Z}$ are generated from the \textit{Aligned Feature}.
To aggregate the style within each semantic region and build their long-range correlation, we introduce a semantic-activation matrix $M$, which can be obtained from the extracted feature of conditional input $X = (x_1,\cdots,x_n)$ by computing its self-attention $M_{ij} = x_{i} \cdot x_{j}$.
As there is only semantic feature in conditional input feature, the semantic-activation matrix accurately measures the self-semantic correlation.
Then the semantic-activation matrix is employed to aggregate the modulation parameters by $\gamma_{\scriptscriptstyle X} = M \cdot \gamma_{\scriptscriptstyle Z}$ and $\mu_{\scriptscriptstyle X} = M \cdot \mu_{\scriptscriptstyle Z}$.
Thus the feature in each position of $\gamma_{\scriptscriptstyle X}$ is determined collectively by a region with the same semantic in $\gamma_{\scriptscriptstyle Z}$ as shown at the bottom of Fig. \ref{im_norm}.
Meanwhile, the long range correlation between modulation parameters with the same semantic is established.

Specially, instead of modulating the generation network directly with these modulation parameters, we first apply the modulation parameters $\gamma_{\scriptscriptstyle X}$ and $\mu_{\scriptscriptstyle X}$ to modulate the activation $X_{act}$ of the conditional input as follows:
\begin{equation}
    X_{act}' = \gamma_{\scriptscriptstyle X} \cdot X_{act} +  \mu_{\scriptscriptstyle X}
\end{equation}
The intuition is that some features cannot be correctly matched if the conditional input contains some parts that do not exist in the exemplar. Thus before injecting the aligned style feature into the generation process, the unmatched feature of conditional input can be effectively corrected according to the accurate semantic information of the conditional input.
Then two sets of modulation parameters $\gamma$ and $\mu$ are further generated from the modulated conditional input $X_{act}'$.

A positional normalization \cite{li2019positional} with variance $\gamma_p$ and mean $\mu_p$ is applied to the activation of the translation network $L_{act}$ to preserve the structure information synthesized in prior layers, followed by a denormalization with $\gamma$ and $\mu$ as follows:
\begin{equation}
    L_{act}' = \gamma  \frac{L_{act} - \mu_p}{\gamma_p} + \mu
\end{equation}

\renewcommand\arraystretch{1.0}
\begin{table*}[t]
\small 
\caption{
Comparing UNITE with state-of-the-art image translation methods: The comparisons were performed over four public datasets with 3 widely used evaluation metrics FID, SWD and LPIPS.
}
\renewcommand\tabcolsep{3.75pt}
\centering 
\begin{tabular}{l||ccc||ccc||ccc||ccc} \hline
\multirow{2}{*}{\textbf{Methods}}
& 
\multicolumn{3}{c||}{\textbf{ADE20K}} & 
\multicolumn{3}{c||}{\textbf{COCO-Stuff}} &
\multicolumn{3}{c||}{\textbf{DeepFashion}} &
\multicolumn{3}{c}{\textbf{CelebA-HQ}}
\\
\cline{2-13}
& FID $\downarrow$ & SWD $\downarrow$ & LPIPS $\uparrow$ & FID $\downarrow$ & SWD  $\downarrow$ & LPIPS  $\uparrow$ & FID  $\downarrow$ & SWD  $\downarrow$ & LPIPS  $\uparrow$ & FID  $\downarrow$ & SWD  $\downarrow$ & LPIPS  $\uparrow$  \\\hline

\textbf{Pix2pixHD\cite{wang2018pix2pixhd}} & 81.80 & 35.70 & N/A    & 121.2 & 44.82 & N/A    & 25.20  & 16.40  & N/A       & 42.70 & 33.30 & N/A     \\

\textbf{Pix2pixSC\cite{wang2019pix2pixsc}} & 56.23 & 24.52 & 0.378    &  77.63  & 26.34 &  0.307    & 28.49  & 21.13  & 0.172    & 49.39 & 33.20 & 0.193      \\

\textbf{StarGAN v2\cite{choi2020starganv2}} & 98.72 & 65.47 & 0.451    &  153.2  & 61.87 &  0.394      & 43.29  & 30.87  & \textbf{0.296}     & 48.63 & 41.96 & 0.214    \\

\textbf{SPADE\cite{park2019spade}} & 33.90 & 19.70 & 0.344     & 49.27 & 19.78 & 0.254   & 36.20  & 27.80  & 0.231  & 31.50 & 26.90 & 0.187       \\

\textbf{SelectionGAN\cite{tang2019selectiongan}}     &  35.10 & 21.82 & 0.382     & 52.41 & 20.32 & 0.277       & 38.31  & 28.21  & 0.223     & 34.67 & 27.34 &  0.191   \\

\textbf{SMIS\cite{zhu2020smis}} & 42.17 &  22.67 &  0.416     & 58.21 &22.65 & 0.311      & 22.23  & 23.73  & 0.240       &  23.71  &  22.23  &  0.201   \\

\textbf{SEAN\cite{zhu2020sean}}     & 24.84 &  10.42  & 0.499       & 37.74 & 16.31 & 0.355      & 16.28  &  17.52  & 0.251       &  18.88    & 19.94 & 0.203      \\

\textbf{CoCosNet\cite{zhang2020cocosnet}}     & 26.40 & 10.50 & 0.560     & 35.23 & 14.54  &  0.391     & 14.40  & 17.20 & 0.272        & 14.30 & 15.30 & 0.208      \\

\hline
\textbf{UNITE}
& \textbf{25.15} & \textbf{10.13} & \textbf{0.571}
& \textbf{33.65} & \textbf{12.18} & \textbf{0.401}
& \textbf{13.08} & \textbf{16.65} & 0.278
& \textbf{13.15} & \textbf{14.91} & \textbf{0.213}
  \\\hline
\end{tabular}
\label{tab_com}
\end{table*}

\renewcommand\arraystretch{1.25}
\begin{table}[t]
\footnotesize
\caption{
Comparing UNITE with state-of-the-art image translation methods over evaluation metrics semantic consistency and style consistency (on dataset ADE20k \cite{zhou2017ade20k}). 
}
\renewcommand\tabcolsep{2pt}
\centering 
\begin{tabular}{l|p{1.3cm}<{\centering}p{1.3cm}<{\centering}|p{1.3cm}<{\centering}p{1.3cm}<{\centering}} \hline
\multirow{2}{*}{\textbf{Methods}} & 
\multicolumn{2}{c|}{\textbf{Semantic Consistency}} & 
\multicolumn{2}{c}{\textbf{Style Consistency}}  
\\
\cline{2-5}
 & ${\rm VGG_{42}}$ $\uparrow$ & ${\rm VGG_{52}}$ $\uparrow$ & ${\rm VGG_{M}}$  $\uparrow$  & ${\rm VGG_{V}}$ $\uparrow$   \\\hline

\textbf{Pix2PixSC \cite{wang2019pix2pixsc}}  &  0.840    & 0.751 &  0.941   &  0.932    \\

\textbf{SPADE \cite{park2019spade}}  &  0.861   & 0.772   & 0.934 &  0.884      \\

\textbf{StarGAN v2 \cite{choi2020starganv2}}  &  0.741  & 0.718   & 0.919 &  0.907      \\

\textbf{SelectionGAN \cite{tang2019selectiongan}}  &  0.843    & 0.785 &  0.951 &  0.912     \\

\textbf{SMIS \cite{zhu2020smis}}  & 0.862     & 0.787 &  0.951  &  0.933     \\

\textbf{SEAN \cite{zhu2020sean}}  &  0.868     & 0.791 &  0.962 &  0.942      \\ 

\textbf{CoCosNet \cite{zhang2020cocosnet}} & 0.878 & 0.790 &  0.986 &  0.965      \\

\hline
\textbf{UNITE}  & \textbf{0.883} & \textbf{0.795} & \textbf{0.990} & \textbf{0.969}     \\\hline
\end{tabular}
\label{tab_consistency}
\end{table}

\subsection{Loss Functions}

The feature transport network and translation network are trained jointly, and will drive each other to achieve better translation.
For clarity purpose, we denote the conditional input and exemplar as $X$ and $Z$, the ground truth as $X'$, the generated image as $Y$,
the feature extractor network for conditional input and exemplar as $F_{X}$ and $F_{Z}$, the translation network as $G$, the discriminator as $D$.

\textbf{Feature Transport Network.}
First, the transported features should be cycle consistent, i.e. the original features should be able to be recovered from the transported features. We thus employ a cycle-consistency loss as follows:
\begin{equation}
    \mathcal{L}_{cyc} = || T^{\top} \cdot T\cdot Z - Z ||_{1}
\end{equation}
where $T$ is the transport plan.
As the two feature extractor networks $F_{X}$ and $F_{Z}$ aim to extract semantic information, the extracted features from the conditional input $X$ and the corresponding ground truth $X'$ should be consistent. A feature consistency loss can thus be defined as follows:
\begin{equation}
    \mathcal{L}_{cst}= || F_{X}(X) - F_{Z}(X') ||_{1}
\end{equation}

\textbf{Translation Network.}
Several losses are employed in the translation network to drive the generation of high-fidelity images.
As the semantic of the generated image should be consistent with the conditional input $X$ or the ground truth $X'$, we employ a perceptual loss $\mathcal{L}_{perc}$ \cite{johnson2016perceptual} to penalize the semantic discrepancy as below:
\begin{equation}
    \mathcal{L}_{perc} = || \phi_{l}(Y) - \phi(X') ||_{1}
\end{equation}
where $\phi_{l}$ represent the activation of layer $l$ in pre-trained VGG-19 \cite{simonyan2014vgg} model.
To ensure the consistency of statistics between the generated image $Y$ and the exemplar $Z$, a contextual loss in \cite{mechrez2018contextual} is adopted as follows:
\begin{equation}
    \mathcal{L}_{cxt} = - \log( \sum_{i} \mathop{\max}\limits_{j} CX_{ij} ( \phi_{l}^{i} (Z), \phi_{l}^{j} (Y) ) ) 
\end{equation}
where $i$ and $j$ are the indexes of the feature map in layer $\phi_{l}$.
Besides, a pseudo pairs loss $\mathcal{L}_{pse}$ as described in \cite{zhang2020cocosnet} is included in training.

The discriminator adopts the same architecture with Patch-GAN  \cite{isola2017pix2pix}. With the adversarial loss $\mathcal{L}_{adv}$, the model can be optimized with the following objective:
\begin{equation}
\begin{split}
    \mathcal{L} = & \mathop{\min}\limits_{F_{X},F_{Z},G} \mathop{\max}\limits_{D} (\lambda_1 \mathcal{L}_{cyc} + \lambda_2 \mathcal{L}_{cst} + \lambda_3 \mathcal{L}_{perc} \\
    &  + \lambda_4 \mathcal{L}_{cxt} + \lambda_5  \mathcal{L}_{pse} + \lambda_6 \mathcal{L}_{adv}) \\
\end{split}
\end{equation}
where the weights $\lambda$ balance the losses in objective.

\section{Experiments}
\subsection{Experimental Settings}

\textbf{Datasets:}
We experiment over multiple public datasets that handle different conditional image translation tasks.

\noindent
$\bullet$ ADE20k \cite{zhou2017ade20k} consists of 20k training images and each image is associated with a 150-class segmentation mask. This is a challenging dataset to most existing methods due to its rich data diversity. We conduct image generation by using its semantic segmentation as conditional inputs.

\noindent
$\bullet$ COCO-Stuff \cite{caesar2018cocostuff} augments COCO \cite{lin2014coco} with pixel-level stuff annotations including 80 thing classes and 91 stuff classes.
We use its layout as conditional inputs. Following \cite{li2019positional}, objects covering less than 2\% of the image are ignored and images with 3 to 8 objects are used in experiments.

\noindent
$\bullet$ CelebA-HQ \cite{liu2015celebahq} consists of 30,000 high quality face images. We use its edge maps as conditional inputs.
The face landmarks are connected as face edges, and the edges in the background are detected by Canny edge detector.

\noindent
$\bullet$ Deepfashion \cite{liu2016deepfashion} contains 52,712 person images with various appearances and poses. 29,000 images are selected as training set and the rest as validation set.
We use its key points as conditional inputs in experiments.

\begin{figure*}[t]
\centering
\includegraphics[width=1.0\linewidth]{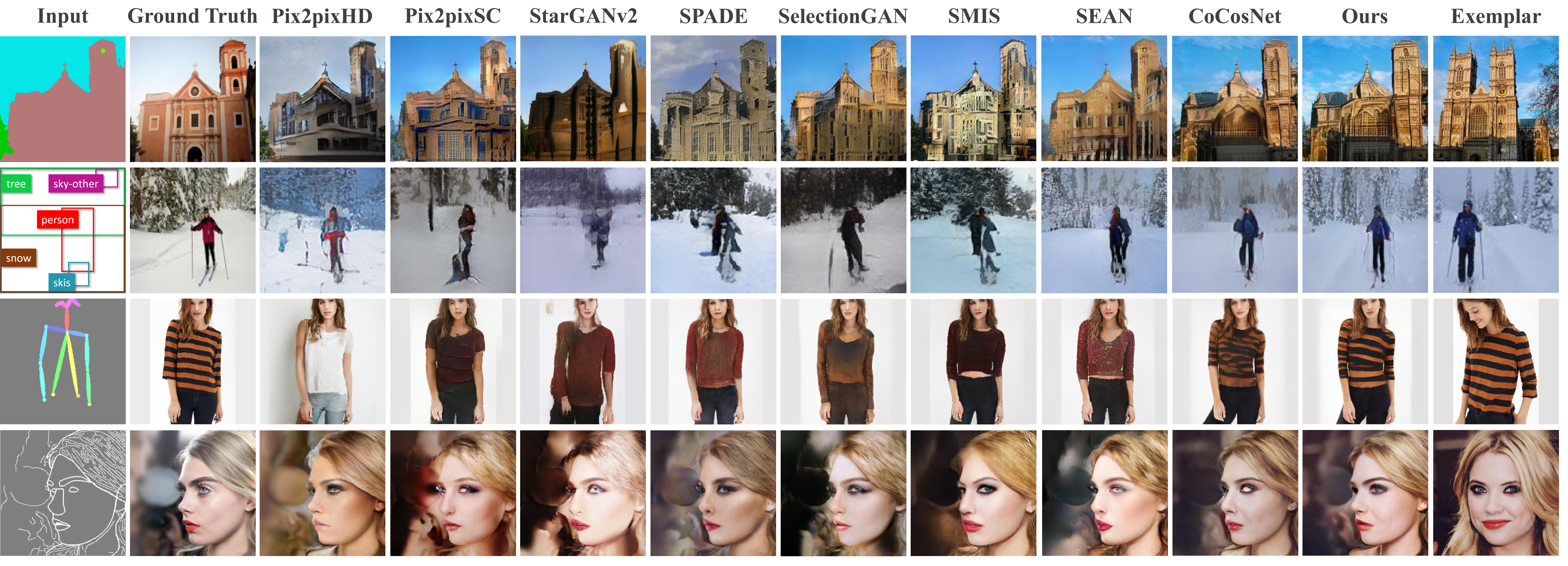}
\caption{
Qualitative illustration of UNITE and state-of-the-art image translation methods over four different types of conditional inputs.
}
\label{im_com}
\end{figure*}

\textbf{Evaluation Metrics:} We adopt several evaluation metrics to assess image translation performance.
\textit{Fr{\'e}chet Inception Score (FID)} \cite{fid} is adopted to measures the distance between the distribution of generated images and real images.
We also adopt \textit{Sliced Wasserstein distance (SWD)} \cite{swd} to measure statistical distance of low level patch distributions.
Besides, \textit{Learned Perceptual Image Patch Similarity (LPIPS)} \cite{zhang2018lpips} is adopted to evaluate the diversity of the translated images with different exemplars, which computes the perceptual distance between image features extracted by AlexNet \cite{alexnet}. 

We also adopt and extend the metrics in \cite{zhang2020cocosnet} to evaluate semantic consistency and style consistency. Specifically, a pre-trained VGG model \cite{simonyan2014vgg} is used to extract high-level features ($relu4\_2$ and $relu5\_2$) of the ground truth and generated images that capture semantic features. The semantic consistency ($\rm VGG_{42}$ and $\rm VGG_{52}$) is defined by the distance between the extracted high-level features as computed by cosine similarity. Similarly, the pre-trained VGG model is applied to extract the low-level feature ($relu1\_2$) of the generated images and exemplars that capture style features. The style consistency ($\rm VGG_{M}$ and $\rm VGG_{V}$) is defined by the distance of channel-wise mean and standard deviation as computed by cosine similarity. 

Besides, we conduct user study (US) to evaluate the images generated under different ablation settings. 100 pairs of generated images were shown to 20 users who select the image with the best visual quality.

\textbf{Implementation Details:}
The learning rate for translation network and discriminator is 1$e$-4 and 4e-4 (the feature transport network is optimized jointly with the translation network).
We use Adam solver with $\beta_{1}=0$ and $\beta_{2}=0.999$. The experiments are conducted on 4 32GB Tesla V100 GPUs with synchronized BatchNorm applied.
The feature size for optimal transport is $64 \times 64$ with feature dimension of 128. The image size is set at $256 \times 256$ for generation tasks using semantic map, edge map, keypoints, and $128\times 128$ for generation task using layout which is consistent with \cite{sun2019lostgan}.

\subsection{Experimental Results} 
We compare UNITE with several state-of-the-art translation methods including 
1) Pix2pixHD \cite{wang2018pix2pixhd}, a supervised image translation method ; 
2) Pix2PixSC \cite{wang2019pix2pixsc}, an example-guided image synthesis model based on Pix2PixHD \cite{wang2018pix2pixhd};
3) StarGAN v2\cite{choi2020starganv2}, a model for multi-modal translation with support for style encoding from reference images;
4) SPADE \cite{park2019spade}, a supervised translation method that supports style injection from an exemplar image;
5) SelectionGAN \cite{tang2019selectiongan}, a guided translation framework with cascaded semantic guidance;
6) SMIS \cite{zhu2020smis}, a network for semantically multi-modal synthesis task with all group convolutions;
7) SEAN \cite{zhu2020sean}, a conditional translation network that can control the style of each individual semantic region;
8) CoCosNet \cite{zhang2020cocosnet}, a leading exemplar-based translation framework that works by building cross-domain correspondences.

\textbf{Quantitative Results:}
In quantitative experiments, all methods synthesize images with the same exemplars except Pix2PixHD \cite{wang2018pix2pixhd} which synthesizes images directly without exemplar guidance (it doesn't support style injection from exemplars). As shown in Table \ref{tab_com}, we compare UNITE with state-of-the-art methods in image quality as measured by FID and SWD and image diversity as measured by LPIPS. We can observe that UNITE outperforms all compared methods over all metrics and tasks consistently. Specifically, UNITE achieves the best FID and SWD which is largely attributed to our designed unbalance optimal transport in accurate feature alignments and semantic-activation normalization in effective style feature injection.
Besides generation quality, UNITE achieves the best generation diversity in LPIPS, thanks to the multi-stage feature transport that aligns features in different scales to faithfully preserve rich textures in exemplars.

Except for high quality and rich diversity, the generated image should preserve consistent semantics with conditional inputs and present consistent styles with exemplars.
Table \ref{tab_consistency} shows the semantic consistency and style consistency evaluated by the metrics described in \textit{Evaluation Metrics}. With our UOT for accurate semantic feature matching and SEACE for effective style injection, UNITE achieves the best semantic consistency and style consistency.

\begin{figure*}[t]
\centering
\includegraphics[width=1.0\linewidth]{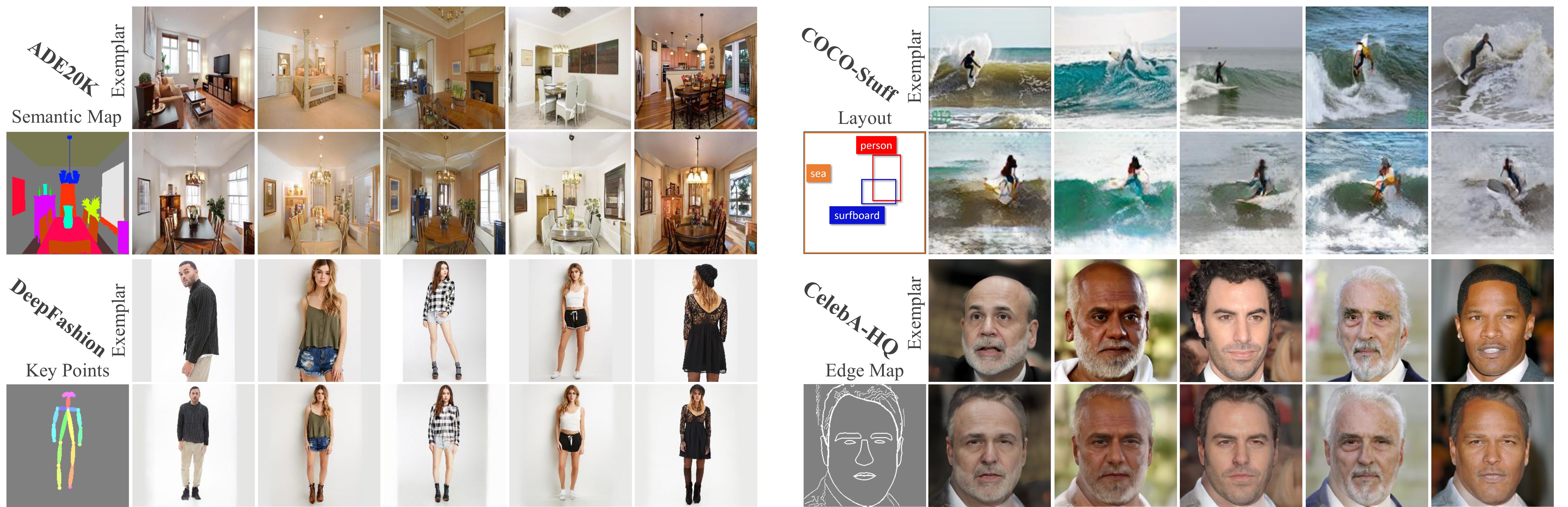}
\caption{
Qualitative illustration of our proposed UNITE with different types of conditional inputs and exemplars.
}
\label{im_diverse}
\end{figure*}

\textbf{Qualitative Evaluation:}
We compare images as generated by different translation methods as shown in Fig. \ref{im_com}. It can be seen that UNITE achieves faithful styles to the exemplars. SPADE \cite{park2019spade}, SMIS \cite{zhu2020smis} and StarGAN v2 \cite{choi2020starganv2} adopt single latent code to encode image styles, which tend to capture global exemplar styles but miss local details. Although SEAN \cite{zhu2020sean} employs multiple latent codes for feature injection, it still struggles to preserve faithful and detailed exemplar style. CoCosNet \cite{zhang2020cocosnet} can preserve certain details, but it adopts cosine similarity to align features which often lead to many-to-one matching and missing details as demonstrated by blurry textures in CoCosNet synthesized images. 
Our UNITE instead adopts UOT to achieve accurate feature alignment and a multi-stage transport to preserve the detailed texture.
Besides, most existing methods tend to produce various artefacts as they do not build long-range dependency between style features. Our UNITE designs SEACE to explicitly build long-range dependency between style features which leads to superior synthesis fidelity as illustrated.

The proposed UNITE also demonstrates superior diversity in image translation as illustrated in Fig. \ref{im_diverse}. We can observe that UNITE is capable of synthesizing various realistic images with faithful style to the given exemplars.

\renewcommand\arraystretch{1.2}
\begin{table}[t]
\small
\caption
{
Ablation studies of our UNITE designs over CelebA-HQ \cite{liu2015celebahq}: The baseline is SPADE that uses spatial denormalization \cite{park2019spade}. COS, OT and UOT mean to include cosine similarity, classical optimal transport and unbalanced optimal transport in feature alignment. SEACE means to use the proposed semantic-activation denormalization to inject style features. MS denotes the multi-stage feature transportation. Model in the last row is the standard UNITE. US denotes the user study metric.
}
\renewcommand\tabcolsep{4.5pt}
\centering 
\begin{tabular}{p{2.5cm}|
p{1cm}<{\centering} 
p{1cm}<{\centering}
p{1.2cm}<{\centering}
p{1cm}<{\centering}} \hline
\textbf{Models} & \textbf{FID}  $\downarrow$ & \textbf{SWD}  $\downarrow$ & \textbf{LPIPS}  $\uparrow$ & \textbf{US}  $\uparrow$
\\
\cline{2-3}
\hline\hline
\textbf{SPADE}      & 31.50  &  26.90  &  0.187   & 0\% \\
\textbf{SPADE+COS}     & 16.32  & 16.10  &  0.201  & 13\% \\
\textbf{SPADE+OT}      & 17.87  & 17.24  &  0.202  & 10\% \\
\textbf{SPADE+UOT}     & 14.02  & 15.41  &  0.206  & 22 \% \\
\textbf{SEACE+UOT}      & 13.46  & 15.12  & 0.208 &  25 \% \\
\hline
\textbf{SEACE+UOT+MS}   & \textbf{13.15} & \textbf{14.91} &  \textbf{0.213} & \textbf{30} \% \\
\hline
\end{tabular}
\label{tab_ablation}
\end{table}

\subsection{Ablation Study}

\begin{figure}[t]
\centering
\includegraphics[width=1.0\linewidth]{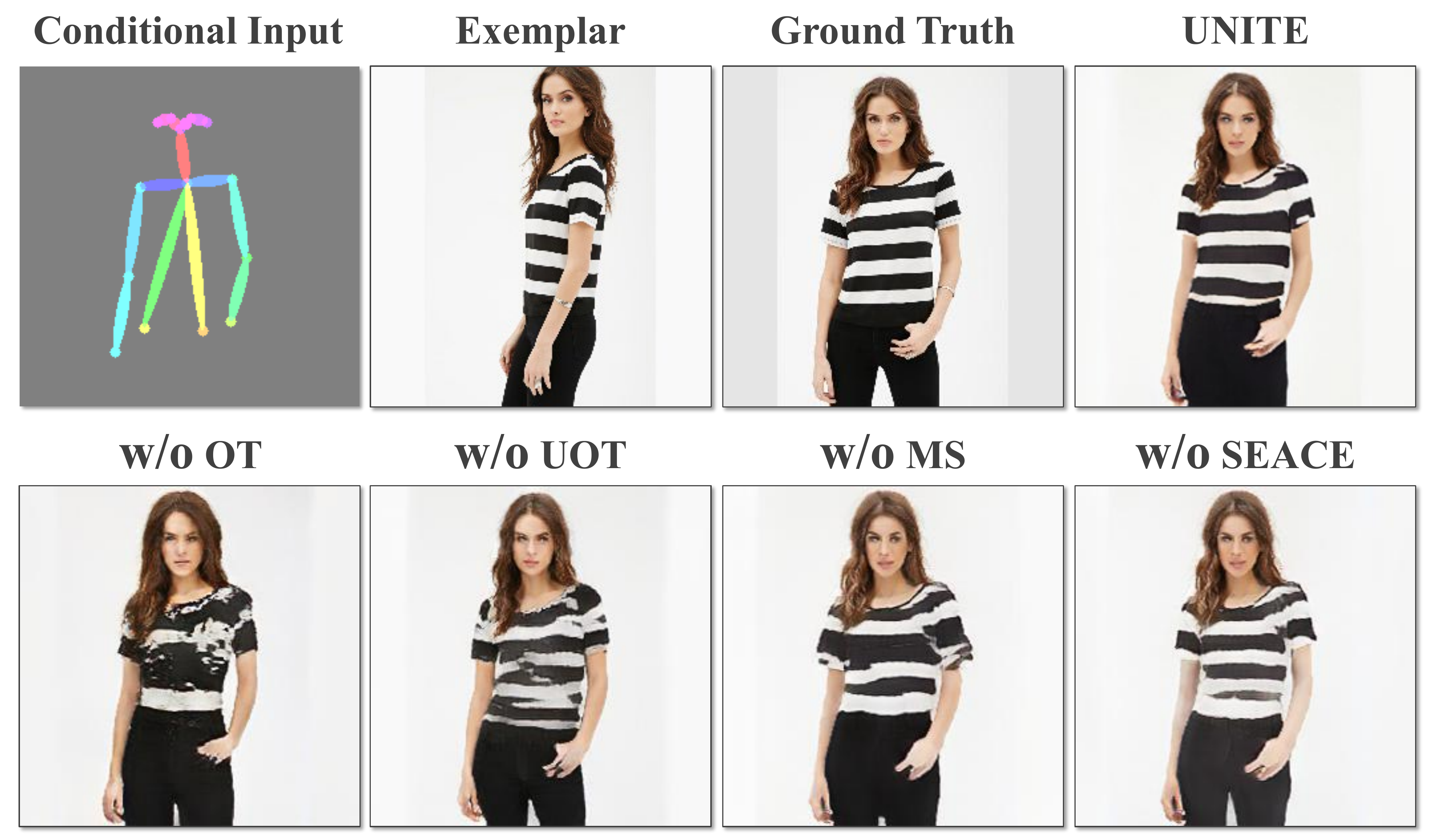}
\caption{
The ablation study of each different design in UNITE as evaluated over a sample from dataset DeepFashion \cite{liu2016deepfashion}. Specially, `w/o OT' denotes image translation without feature alignment, `w/o UOT' denotes using classical OT (without learnt unbalanced weights) to align features.
}
\label{im_ablation}
\end{figure}

We conduct extensive ablation studies over CelebA-HQ \cite{liu2015celebahq} to validate the effectiveness of our designs. As Table \ref{tab_ablation} shows, SPADE \cite{park2019spade} is the baseline which achieves image translation directly without feature alignment. When cosine similarity is included to align features, the translation is improved significantly. While replacing cosine similarity with classical optimal transport, the performance does is clearly aggravated as classical optimal transport introduce many false matchings. However, the translation performance improves clearly when our UOT is included, largely attributed to that UOT adaptively learns the feature masses and suppresses false and many-to-one matching effectively. When replacing SPADE with our proposed SEACE, the FID score is improved clearly by 0.73. Additionally, the SWD and LPIP scores are improved clearly when our proposed multi-stage feature transport is included. We also performed qualitative ablation studies on DeepFashion \cite{liu2016deepfashion} by removing each of our designs from the complete UNITE model.
As Fig. \ref{im_ablation} shows, our designed UOT, MS and SEACE all contribute to the high-fidelity realistic image translation clearly.

\section{Conclusions}
This paper presents UNITE, an exemplar-based image translation framework that adopts unbalanced optimal transport to align the feature between conditional input and exemplar, which effectively transport the style of the exemplar to the conditional input. A multi-stage feature transport manner is applied to preserved more detailed deep features. 
To inject aligned the style feature into the generation process, we propose a novel semantic-activation normalization which builds the semantic coherence between style features with the same semantic in style injection.
Quantitative and qualitative experiments show that UNITE is capable of generating high-fidelity images with consistent semantic with the conditional input and faithful style to the exemplar.

{\small
\bibliographystyle{ieee_fullname}
\bibliography{egbib}

\begin{thebibliography}{10}\itemsep=-1pt

\bibitem{caesar2018cocostuff}
Holger Caesar, Jasper Uijlings, and Vittorio Ferrari.
\newblock Coco-stuff: Thing and stuff classes in context.
\newblock In {\em Proceedings of the IEEE Conference on Computer Vision and
  Pattern Recognition}, pages 1209--1218, 2018.

\bibitem{chizat2016uot}
Lenaic Chizat, Gabriel Peyré, Bernhard Schmitzer, and François-Xavier
  Vialard.
\newblock Scaling algorithms for unbalanced transport problems.
\newblock In {\em arXiv:1607.05816}, 2016.

\bibitem{choi2020starganv2}
Yunjey Choi, Youngjung Uh, Jaejun Yoo, and Jung-Woo Ha.
\newblock Stargan v2: Diverse image synthesis for multiple domains.
\newblock In {\em Proceedings of the IEEE/CVF Conference on Computer Vision and
  Pattern Recognition}, pages 8188--8197, 2020.

\bibitem{courty2016optimal}
Nicolas Courty, R{\'e}mi Flamary, Devis Tuia, and Alain Rakotomamonjy.
\newblock Optimal transport for domain adaptation.
\newblock {\em IEEE transactions on pattern analysis and machine intelligence},
  39(9):1853--1865, 2016.

\bibitem{cuturi2013sinkhorn}
Marco Cuturi.
\newblock Sinkhorn distances: Lightspeed computation of optimal transport.
\newblock In {\em Advances in neural information processing systems}, pages
  2292--2300, 2013.

\bibitem{he2018colorization}
Mingming He, Dongdong Chen, Jing Liao, Pedro~V Sander, and Lu Yuan.
\newblock Deep exemplar-based colorization.
\newblock {\em ACM Transactions on Graphics (TOG)}, 37(4):1--16, 2018.

\bibitem{fid}
Martin Heusel, Hubert Ramsauer, Thomas Unterthiner, Bernhard Nessler, and Sepp
  Hochreiter.
\newblock Gans trained by a two time-scale update rule converge to a local nash
  equilibrium.
\newblock In {\em Advances in neural information processing systems}, pages
  6626--6637, 2017.

\bibitem{huang2017adain}
Xun Huang and Serge Belongie.
\newblock Arbitrary style transfer in real-time with adaptive instance
  normalization.
\newblock In {\em Proceedings of the IEEE International Conference on Computer
  Vision}, pages 1501--1510, 2017.

\bibitem{huang2018multimodal}
Xun Huang, Ming-Yu Liu, Serge Belongie, and Jan Kautz.
\newblock Multimodal unsupervised image-to-image translation.
\newblock In {\em Proceedings of the European Conference on Computer Vision
  (ECCV)}, pages 172--189, 2018.

\bibitem{isola2017pix2pix}
Phillip Isola, Jun-Yan Zhu, Tinghui Zhou, and Alexei~A Efros.
\newblock Image-to-image translation with conditional adversarial networks.
\newblock In {\em Proceedings of the IEEE conference on computer vision and
  pattern recognition}, pages 1125--1134, 2017.

\bibitem{johnson2016perceptual}
Justin Johnson, Alexandre Alahi, and Li Fei-Fei.
\newblock Perceptual losses for real-time style transfer and super-resolution.
\newblock In {\em European conference on computer vision}, pages 694--711.
  Springer, 2016.

\bibitem{swd}
Tero Karras, Timo Aila, Samuli Laine, and Jaakko Lehtinen.
\newblock Progressive growing of gans for improved quality, stability, and
  variation.
\newblock {\em arXiv preprint arXiv:1710.10196}, 2017.

\bibitem{kingma2013vae}
Diederik~P Kingma and Max Welling.
\newblock Auto-encoding variational bayes.
\newblock {\em arXiv preprint arXiv:1312.6114}, 2013.

\bibitem{knight2008sinkhorn}
Philip~A Knight.
\newblock The sinkhorn--knopp algorithm: convergence and applications.
\newblock {\em SIAM Journal on Matrix Analysis and Applications},
  30(1):261--275, 2008.

\bibitem{koksal2020rf}
Ali Koksal and Shijian Lu.
\newblock Rf-gan: A light and reconfigurable network for unpaired
  image-to-image translation.
\newblock In {\em Proceedings of the Asian Conference on Computer Vision},
  2020.

\bibitem{kolkin2019style}
Nicholas Kolkin, Jason Salavon, and Gregory Shakhnarovich.
\newblock Style transfer by relaxed optimal transport and self-similarity.
\newblock In {\em Proceedings of the IEEE Conference on Computer Vision and
  Pattern Recognition}, pages 10051--10060, 2019.

\bibitem{alexnet}
Alex Krizhevsky, Ilya Sutskever, and Geoffrey~E Hinton.
\newblock Imagenet classification with deep convolutional neural networks.
\newblock In {\em NIPS}, 2012.

\bibitem{lee2018diverse}
Hsin-Ying Lee, Hung-Yu Tseng, Jia-Bin Huang, Maneesh Singh, and Ming-Hsuan
  Yang.
\newblock Diverse image-to-image translation via disentangled representations.
\newblock In {\em Proceedings of the European conference on computer vision
  (ECCV)}, pages 35--51, 2018.

\bibitem{li2019positional}
Boyi Li, Felix Wu, Kilian~Q Weinberger, and Serge Belongie.
\newblock Positional normalization.
\newblock In {\em Advances in Neural Information Processing Systems}, pages
  1622--1634, 2019.

\bibitem{li2020bachgan}
Yandong Li, Yu Cheng, Zhe Gan, Licheng Yu, Liqiang Wang, and Jingjing Liu.
\newblock Bachgan: High-resolution image synthesis from salient object layout.
\newblock In {\em Proceedings of the IEEE Conference on Computer Vision and
  Pattern Recognition}, 2020.

\bibitem{liero2015uot}
Matthias Liero, Alexander Mielke, and Giuseppe Savaré.
\newblock Optimal entropy-transport problems and a new hellinger-kantorovich
  distance between positive measures.
\newblock {\em arXiv:1508.07941}, 2015.

\bibitem{lin2014coco}
Tsung-Yi Lin, Michael Maire, Serge Belongie, James Hays, Pietro Perona, Deva
  Ramanan, Piotr Doll{\'a}r, and C~Lawrence Zitnick.
\newblock Microsoft coco: Common objects in context.
\newblock In {\em European conference on computer vision}, pages 740--755.
  Springer, 2014.

\bibitem{liu2020scot}
Yanbin Liu, Linchao Zhu, Makoto Yamada, and Yi Yang.
\newblock Semantic correspondence as an optimal transport problem.
\newblock In {\em Proceedings of the IEEE/CVF Conference on Computer Vision and
  Pattern Recognition}, pages 4463--4472, 2020.

\bibitem{liu2016deepfashion}
Ziwei Liu, Ping Luo, Shi Qiu, Xiaogang Wang, and Xiaoou Tang.
\newblock Deepfashion: Powering robust clothes recognition and retrieval with
  rich annotations.
\newblock In {\em Proceedings of the IEEE conference on computer vision and
  pattern recognition}, pages 1096--1104, 2016.

\bibitem{liu2015celebahq}
Ziwei Liu, Ping Luo, Xiaogang Wang, and Xiaoou Tang.
\newblock Deep learning face attributes in the wild.
\newblock In {\em Proceedings of the IEEE international conference on computer
  vision}, pages 3730--3738, 2015.

\bibitem{ma2018exemplar}
Liqian Ma, Xu Jia, Stamatios Georgoulis, Tinne Tuytelaars, and Luc Van~Gool.
\newblock Exemplar guided unsupervised image-to-image translation with semantic
  consistency.
\newblock In {\em International Conference on Learning Representations}, 2018.

\bibitem{ma2017pose}
Liqian Ma, Xu Jia, Qianru Sun, Bernt Schiele, Tinne Tuytelaars, and Luc
  Van~Gool.
\newblock Pose guided person image generation.
\newblock In {\em Advances in neural information processing systems}, pages
  406--416, 2017.

\bibitem{mechrez2018contextual}
Roey Mechrez, Itamar Talmi, and Lihi Zelnik-Manor.
\newblock The contextual loss for image transformation with non-aligned data.
\newblock In {\em Proceedings of the European Conference on Computer Vision
  (ECCV)}, pages 768--783, 2018.

\bibitem{men2020adgan}
Yifang Men, Yiming Mao, Yuning Jiang, Wei-Ying Ma, and Zhouhui Lian.
\newblock Controllable person image synthesis with attribute-decomposed gan.
\newblock In {\em Proceedings of the IEEE/CVF Conference on Computer Vision and
  Pattern Recognition}, pages 5084--5093, 2020.

\bibitem{park2019spade}
Taesung Park, Ming-Yu Liu, Ting-Chun Wang, and Jun-Yan Zhu.
\newblock Semantic image synthesis with spatially-adaptive normalization.
\newblock In {\em Proceedings of the IEEE Conference on Computer Vision and
  Pattern Recognition}, pages 2337--2346, 2019.

\bibitem{peyre2019computational}
Gabriel Peyr{\'e}, Marco Cuturi, et~al.
\newblock Computational optimal transport: With applications to data science.
\newblock {\em Foundations and Trends{\textregistered} in Machine Learning},
  11(5-6):355--607, 2019.

\bibitem{shrivastava2017simgan}
Ashish Shrivastava, Tomas Pfister, Oncel Tuzel, Joshua Susskind, Wenda Wang,
  and Russell Webb.
\newblock Learning from simulated and unsupervised images through adversarial
  training.
\newblock In {\em Proceedings of the IEEE conference on computer vision and
  pattern recognition}, pages 2107--2116, 2017.

\bibitem{simonyan2014vgg}
Karen Simonyan and Andrew Zisserman.
\newblock Very deep convolutional networks for large-scale image recognition.
\newblock {\em arXiv preprint arXiv:1409.1556}, 2014.

\bibitem{sinkhorn1967concerning}
Richard Sinkhorn and Paul Knopp.
\newblock Concerning nonnegative matrices and doubly stochastic matrices.
\newblock {\em Pacific Journal of Mathematics}, 21(2):343--348, 1967.

\bibitem{sun2019lostgan}
Wei Sun and Tianfu Wu.
\newblock Image synthesis from reconfigurable layout and style.
\newblock In {\em Proceedings of the IEEE International Conference on Computer
  Vision}, pages 10531--10540, 2019.

\bibitem{tang2019cycle}
Hao Tang, Dan Xu, Gaowen Liu, Wei Wang, Nicu Sebe, and Yan Yan.
\newblock Cycle in cycle generative adversarial networks for keypoint-guided
  image generation.
\newblock In {\em Proceedings of the 27th ACM International Conference on
  Multimedia}, pages 2052--2060, 2019.

\bibitem{tang2019selectiongan}
Hao Tang, Dan Xu, Nicu Sebe, Yanzhi Wang, Jason~J Corso, and Yan Yan.
\newblock Multi-channel attention selection gan with cascaded semantic guidance
  for cross-view image translation.
\newblock In {\em Proceedings of the IEEE Conference on Computer Vision and
  Pattern Recognition}, pages 2417--2426, 2019.

\bibitem{villani2008ot}
C{\'e}dric Villani.
\newblock {\em Optimal transport: old and new}, volume 338.
\newblock Springer Science \& Business Media, 2008.

\bibitem{wang2019pix2pixsc}
Miao Wang, Guo-Ye Yang, Ruilong Li, Run-Ze Liang, Song-Hai Zhang, Peter~M Hall,
  and Shi-Min Hu.
\newblock Example-guided style-consistent image synthesis from semantic
  labeling.
\newblock In {\em Proceedings of the IEEE Conference on Computer Vision and
  Pattern Recognition}, pages 1495--1504, 2019.

\bibitem{wang2018pix2pixhd}
Ting-Chun Wang, Ming-Yu Liu, Jun-Yan Zhu, Andrew Tao, Jan Kautz, and Bryan
  Catanzaro.
\newblock High-resolution image synthesis and semantic manipulation with
  conditional gans.
\newblock In {\em Proceedings of the IEEE conference on computer vision and
  pattern recognition}, pages 8798--8807, 2018.

\bibitem{wu2020leed}
Rongliang Wu and Shijian Lu.
\newblock Leed: Label-free expression editing via disentanglement.
\newblock In {\em European Conference on Computer Vision}, pages 781--798.
  Springer, 2020.

\bibitem{wu2020cascade}
Rongliang Wu, Gongjie Zhang, Shijian Lu, and Tao Chen.
\newblock Cascade ef-gan: Progressive facial expression editing with local
  focuses.
\newblock In {\em Proceedings of the IEEE/CVF Conference on Computer Vision and
  Pattern Recognition}, pages 5021--5030, 2020.

\bibitem{xue2018accurate}
Chuhui Xue, Shijian Lu, and Fangneng Zhan.
\newblock Accurate scene text detection through border semantics awareness and
  bootstrapping.
\newblock In {\em Proceedings of the European conference on computer vision
  (ECCV)}, pages 355--372, 2018.

\bibitem{yu2018inpainting}
Jiahui Yu, Zhe Lin, Jimei Yang, Xiaohui Shen, Xin Lu, and Thomas~S Huang.
\newblock Generative image inpainting with contextual attention.
\newblock In {\em Proceedings of the IEEE conference on computer vision and
  pattern recognition}, pages 5505--5514, 2018.

\bibitem{yu2021diverse}
Yingchen Yu, Fangneng Zhan, Rongliang Wu, Jianxiong Pan, Kaiwen Cui, Shijian
  Lu, Feiying Ma, Xuansong Xie, and Chunyan Miao.
\newblock Diverse image inpainting with bidirectional and autoregressive
  transformers.
\newblock {\em arXiv preprint arXiv:2104.12335}, 2021.

\bibitem{zhan2019acgan}
Fangneng Zhan, Jiaxing Huang, and Shijian Lu.
\newblock Adaptive composition gan towards realistic image synthesis.
\newblock {\em arXiv preprint arXiv:1905.04693}, 2019.

\bibitem{zhan2019esir}
Fangneng Zhan and Shijian Lu.
\newblock Esir: End-to-end scene text recognition via iterative image
  rectification.
\newblock In {\em Proceedings of the IEEE Conference on Computer Vision and
  Pattern Recognition}, pages 2059--2068, 2019.

\bibitem{zhan2018verisimilar}
Fangneng Zhan, Shijian Lu, and Chuhui Xue.
\newblock Verisimilar image synthesis for accurate detection and recognition of
  texts in scenes.
\newblock In {\em Proceedings of the European Conference on Computer Vision
  (ECCV)}, pages 249--266, 2018.

\bibitem{zhan2020aicnet}
Fangneng Zhan, Shijian Lu, Changgong Zhang, Feiying Ma, and Xuansong Xie.
\newblock Adversarial image composition with auxiliary illumination.
\newblock In {\em Proceedings of the Asian Conference on Computer Vision},
  2020.

\bibitem{zhan2020towards}
Fangneng Zhan, Shijian Lu, Changgong Zhang, Feiying Ma, and Xuansong Xie.
\newblock Towards realistic 3d embedding via view alignment.
\newblock {\em arXiv preprint arXiv:2007.07066}, 2020.

\bibitem{zhan2019gadan}
Fangneng Zhan, Chuhui Xue, and Shijian Lu.
\newblock Ga-dan: Geometry-aware domain adaptation network for scene text
  detection and recognition.
\newblock In {\em Proceedings of the IEEE International Conference on Computer
  Vision}, pages 9105--9115, 2019.

\bibitem{zhan2021rabit}
Fangneng Zhan, Yingchen Yu, Rongliang Wu, Kaiwen Cui, Aoran Xiao, Shijian Lu,
  and Ling Shao.
\newblock Bi-level feature alignment for semantic image translation \&
  manipulation.
\newblock {\em arXiv preprint}, 2021.

\bibitem{zhan2021gmlight}
Fangneng Zhan, Yingchen Yu, Rongliang Wu, Changgong Zhang, Shijian Lu, Ling
  Shao, Feiying Ma, and Xuansong Xie.
\newblock Gmlight: Lighting estimation via geometric distribution
  approximation.
\newblock {\em arXiv preprint arXiv:2102.10244}, 2021.

\bibitem{zhan2020sagan}
Fangneng Zhan and Changgong Zhang.
\newblock Spatial-aware gan for unsupervised person re-identification.
\newblock {\em Proceedings of the International Conference on Pattern
  Recognition}, 2020.

\bibitem{zhan2020emlight}
Fangneng Zhan, Changgong Zhang, Yingchen Yu, Yuan Chang, Shijian Lu, Feiying
  Ma, and Xuansong Xie.
\newblock Emlight: Lighting estimation via spherical distribution
  approximation.
\newblock {\em arXiv preprint arXiv:2012.11116}, 2020.

\bibitem{zhan2019scene}
Fangneng Zhan, Hongyuan Zhu, and Shijian Lu.
\newblock Scene text synthesis for efficient and effective deep network
  training.
\newblock {\em arXiv preprint arXiv:1901.09193}, 2019.

\bibitem{zhan2019sfgan}
Fangneng Zhan, Hongyuan Zhu, and Shijian Lu.
\newblock Spatial fusion gan for image synthesis.
\newblock In {\em Proceedings of the IEEE conference on computer vision and
  pattern recognition}, pages 3653--3662, 2019.

\bibitem{zhang2019colorization}
Bo Zhang, Mingming He, Jing Liao, Pedro~V Sander, Lu Yuan, Amine Bermak, and
  Dong Chen.
\newblock Deep exemplar-based video colorization.
\newblock In {\em Proceedings of the IEEE Conference on Computer Vision and
  Pattern Recognition}, pages 8052--8061, 2019.

\bibitem{zhang2021deep}
Changgong Zhang, Fangneng Zhan, and Yuan Chang.
\newblock Deep monocular 3d human pose estimation via cascaded
  dimension-lifting.
\newblock {\em arXiv preprint arXiv:2104.03520}, 2021.

\bibitem{zhang2021defect}
Gongjie Zhang, Kaiwen Cui, Tzu-Yi Hung, and Shijian Lu.
\newblock Defect-gan: High-fidelity defect synthesis for automated defect
  inspection.
\newblock In {\em Proceedings of the IEEE/CVF Winter Conference on Applications
  of Computer Vision}, pages 2524--2534, 2021.

\bibitem{zhang2019self}
Han Zhang, Ian Goodfellow, Dimitris Metaxas, and Augustus Odena.
\newblock Self-attention generative adversarial networks.
\newblock In {\em International Conference on Machine Learning}, pages
  7354--7363. PMLR, 2019.

\bibitem{zhang2020cocosnet}
Pan Zhang, Bo Zhang, Dong Chen, Lu Yuan, and Fang Wen.
\newblock Cross-domain correspondence learning for exemplar-based image
  translation.
\newblock In {\em Proceedings of the IEEE/CVF Conference on Computer Vision and
  Pattern Recognition}, pages 5143--5153, 2020.

\bibitem{zhang2018lpips}
Richard Zhang, Phillip Isola, Alexei~A Efros, Eli Shechtman, and Oliver Wang.
\newblock The unreasonable effectiveness of deep features as a perceptual
  metric.
\newblock In {\em Proceedings of the IEEE conference on computer vision and
  pattern recognition}, pages 586--595, 2018.

\bibitem{zhao2019layout2im}
Bo Zhao, Lili Meng, Weidong Yin, and Leonid Sigal.
\newblock Image generation from layout.
\newblock In {\em Proceedings of the IEEE Conference on Computer Vision and
  Pattern Recognition}, pages 8584--8593, 2019.

\bibitem{zhou2017ade20k}
Bolei Zhou, Hang Zhao, Xavier Puig, Sanja Fidler, Adela Barriuso, and Antonio
  Torralba.
\newblock Scene parsing through ade20k dataset.
\newblock In {\em Proceedings of the IEEE conference on computer vision and
  pattern recognition}, pages 633--641, 2017.

\bibitem{zhu2017toward}
Jun-Yan Zhu, Richard Zhang, Deepak Pathak, Trevor Darrell, Alexei~A Efros,
  Oliver Wang, and Eli Shechtman.
\newblock Toward multimodal image-to-image translation.
\newblock In {\em Advances in neural information processing systems}, pages
  465--476, 2017.

\bibitem{zhu2020sean}
Peihao Zhu, Rameen Abdal, Yipeng Qin, and Peter Wonka.
\newblock Sean: Image synthesis with semantic region-adaptive normalization.
\newblock In {\em Proceedings of the IEEE/CVF Conference on Computer Vision and
  Pattern Recognition}, pages 5104--5113, 2020.

\bibitem{zhu2020smis}
Zhen Zhu, Zhiliang Xu, Ansheng You, and Xiang Bai.
\newblock Semantically multi-modal image synthesis.
\newblock In {\em Proceedings of the IEEE/CVF Conference on Computer Vision and
  Pattern Recognition}, pages 5467--5476, 2020.

\end{thebibliography}
}

\end{document}